\documentclass[twocolumn]{article}
\usepackage{mathtools}
\usepackage{amsmath}
\usepackage{amssymb}
\usepackage{amsfonts}
\usepackage[T1]{fontenc}
\usepackage{arxiv}

\usepackage{url}
\usepackage{booktabs}
\usepackage{amsfonts}
\usepackage{nicefrac}
\usepackage{microtype}
\usepackage{lipsum}
\usepackage{graphicx}
\usepackage{doi}
\usepackage{titlesec}
\usepackage{setspace}
\usepackage{xcolor}
\usepackage{amsmath}
\usepackage{tabularx}
\usepackage{colortbl}

\usepackage{adjustbox}
\usepackage{libertinus}
\usepackage[backend=biber,style=ieee]{biblatex}
\usepackage{caption}
\title{\large\bfseries\textit{GSplatLoc} : Ultra-Precise Camera
Localization via 3D Gaussian Splatting}

\usepackage{titling}

\usepackage{hyperref}

\hypersetup{
    pdftitle={GSplatLoc : Ultra-Precise Camera Localization via 3D
Gaussian Splatting},
    pdfauthor={Atticus J. Zeller},
    colorlinks=true,
    linkcolor=blue,
    citecolor=green,
    urlcolor=cyan,
}
\newcommand{\authorinfo}[4]{%
  \begin{tabular}[t]{c}
    \href{#1}{\includegraphics[scale=0.06]{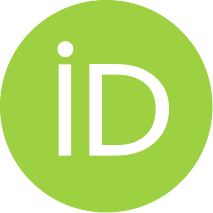}\hspace{1mm}\textbf{#2}}#4\\
    #3\\
  \end{tabular}%
}
\renewcommand{\and}{\hspace{2em}}

\author{%
\authorinfo{https://orcid.org/0009-0008-5460-325X}{Atticus J.
Zeller}{Southeast University Chengxian College\\Nanjing, China}{}\and%
\authorinfo{}{Haijuan Wu}{Southeast University Chengxian College\\Nanjing, China}{}%
}

\begin{document}

\twocolumn[
\begin{@twocolumnfalse}
  \maketitle
  \begin{abstract}
    We present \textbf{GSplatLoc}, a camera localization method that
    leverages the differentiable rendering capabilities of 3D Gaussian
    splatting for ultra-precise pose estimation. By formulating pose
    estimation as a gradient-based optimization problem that minimizes
    discrepancies between rendered depth maps from a pre-existing 3D
    Gaussian scene and observed depth images, GSplatLoc achieves
    translational errors within \textbf{0.01\,cm} and near-zero
    rotational errors on the Replica dataset---significantly
    outperforming existing methods. Evaluations on the Replica and TUM
    RGB-D datasets demonstrate the method's robustness in challenging
    indoor environments with complex camera motions. GSplatLoc sets a
    new benchmark for localization in dense mapping, with important
    implications for applications requiring accurate real-time
    localization, such as robotics and augmented reality. Code is
    available at \url{https://github.com/AtticusZeller/GsplatLoc}.
  \end{abstract}
  \vspace{1cm}
\end{@twocolumnfalse}
]

\section{Introduction}\label{introduction}

Visual localization\autocite{scaramuzzaVisualOdometryTutorial2011},
specifically the task of estimating camera position and orientation
(pose estimation) for a given image within a known scene, is a
fundamental challenge in computer vision. Accurate pose estimation is
crucial for applications like autonomous robotics (e.g., self-driving
cars), as well as Augmented and Virtual Reality systems. Although Visual
Simultaneous Localization and Mapping (Visual
SLAM)\autocite{durrant-whyteSimultaneousLocalizationMapping2006,davisonMonoSLAMRealtimeSingle2007}combines
both mapping and pose estimation, this paper focuses specifically on the
localization component, which is essential for real-time tracking in
dynamic environments.

Traditional SLAM systems \autocite{kerlDenseVisualSLAM2013} have
demonstrated accurate pose estimation across diverse environments.
However, their underlying 3D representations (e.g., point clouds,
meshes, and surfels) exhibit
limitations\autocite{newcombeKinectfusionRealtimeDense2011,rusinkiewiczEfficientVariantsICP2001}
in flexibility for tasks like photorealistic scene exploration and
fine-grained map updates. Recent methods utilizing Neural Radiance
Fields (NeRF) \autocite{mildenhallNeRFRepresentingScenes2022} for
surface reconstruction and view rendering have inspired novel SLAM
approaches \autocite{sandstromPointslamDenseNeural2023}, which show
promising\autocite{sucarImapImplicitMapping2021,zhuNiceslamNeuralImplicit2022}
results in tracking and scene modeling. Despite these
advances\autocite{garbinFastnerfHighfidelityNeural2021}, existing
NeRF-based methods rely on computationally expensive volume rendering
pipelines, limiting their ability to perform real-time \textbf{pose
estimation} effectively.

The development of \textbf{3D Gaussian Splatting}
\autocite{kerbl3DGaussianSplatting2023} for efficient novel view
synthesis presents a promising solution to these limitations. Its
rasterization-based rendering pipeline enables faster image-level
rendering, making it more suitable for real-time applications. However,
integrating 3D Gaussian fields into SLAM systems still faces challenges,
such as overfitting to input images due to anisotropic Gaussian fields
and a lack of explicit multi-view constraints.

Current SLAM methods using 3D Gaussian Splatting, such as RTG-SLAM
\autocite{pengRTGSLAMRealtime3D2024} and GS-ICP-SLAM
\autocite{haRGBDGSICPSLAM2024}, rely primarily on ICP-based techniques
for pose estimation. Other approaches, like Gaussian-SLAM
\autocite{yugayGaussianSLAMPhotorealisticDense2024}, adapt traditional
RGB-D odometry methods. While these methods have shown potential, they
often do not fully exploit the differentiable nature of the Gaussian
Splatting representation, particularly for real-time and efficient
\textbf{pose estimation}.

In this paper, we introduce \textbf{GSplatLoc}, a novel camera
localization method that leverages the differentiable properties of 3D
Gaussian Splatting for efficient and accurate pose estimation. By
focusing solely on the localization aspect rather than the full SLAM
pipeline, GSplatLoc allows for more efficient utilization of the scene
representation and camera pose estimation, seamlessly integrating into
existing Gaussian Splatting SLAM frameworks or other deep learning tasks
focused on localization.

Our main contributions include presenting a GPU-accelerated framework
for real-time camera localization, based on a comprehensive theoretical
analysis of camera pose derivatives in 3D Gaussian Splatting; proposing
a novel optimization approach that fully exploits the differentiable
nature of the rendering process for camera pose estimation given a 3D
Gaussian scene; and demonstrating the effectiveness of our method
through extensive experiments, showing competitive or superior pose
estimation results compared to state-of-the-art SLAM approaches
utilizing advanced scene representations. By specifically addressing the
challenges of localization in Gaussian Splatting-based scenes, GSplatLoc
opens new avenues for high-precision camera pose estimation in complex
environments, contributing to the ongoing advancement of visual
localization systems and pushing the boundaries of accuracy and
real-time performance in 3D scene understanding and navigation.

\section{Related Work}\label{related-work}

Camera localization is a fundamental problem in computer vision and
robotics, crucial for applications such as autonomous navigation,
augmented reality, and 3D reconstruction. Accurate and efficient pose
estimation remains challenging, especially in complex and dynamic
environments. In this section, we review the evolution of camera
localization methodologies, focusing on classical RGB-D localization
methods, NeRF-based approaches, and recent advancements in
Gaussian-based techniques that utilize Iterative Closest Point (ICP)
algorithms. We highlight their contributions and limitations, which
motivate the development of our proposed method.

\subsection{Classical RGB-D
Localization}\label{classical-rgb-d-localization}

Traditional RGB-D localization methods leverage both color and depth
information to estimate camera poses. These methods can be broadly
categorized into feature-based, direct, and hybrid approaches.

\textbf{Feature-Based Methods} involve extracting and matching keypoints
across frames to estimate camera motion. Notable systems such as
ORB-SLAM2 \autocite{mur-artalOrbslam2OpensourceSlam2017} , ORB-SLAM3
\autocite{camposOrbslam3AccurateOpensource2021} and
\autocite{gauglitzEvaluationInterestPoint2011} rely on sparse feature
descriptors like ORB features. These systems have demonstrated robust
performance in various environments, benefiting from the maturity of
feature detection and matching algorithms. However, their reliance on
distinct visual features makes them less effective in textureless or
repetitive scenes. While they can utilize depth information for scale
estimation and map refinement, they primarily depend on RGB data for
pose estimation, making them susceptible to lighting changes and
appearance variations.

\textbf{Direct Methods}\autocite{engelDirectSparseOdometry2017} estimate
camera motion by minimizing the photometric error between consecutive
frames, utilizing all available pixel information. Methods such as Dense
Visual Odometry (DVO)
\autocite{kerlDenseVisualSLAM2013,kerlRobustOdometryEstimation2013} and
DTAM\autocite{newcombeDTAMDenseTracking2011} incorporate depth data to
enhance pose estimation accuracy. These methods can achieve high
precision in well-lit, textured environments but are sensitive to
illumination changes and require good initialization to avoid local
minima. The computational cost of processing all pixel data poses
challenges for real-time applications. Additionally, the reliance on
photometric consistency makes them vulnerable to lighting variations and
dynamic scenes.

\textbf{Hybrid Approaches} combine the strengths of feature-based and
direct methods. ElasticFusion
\autocite{whelanElasticFusionRealtimeDense2016} integrates surfel-based
mapping with real-time camera tracking, using both photometric and
geometric information. DVO-SLAM
\autocite{kerlRobustOdometryEstimation2013} combines geometric and
photometric alignment for improved robustness. However, these methods
often involve complex pipelines and can be computationally intensive due
to dense map representations and intricate data association processes.

Despite their successes, classical methods face challenges in balancing
computational efficiency with pose estimation accuracy, particularly in
dynamic or low-texture environments. They may not fully exploit the
potential of depth information for robust pose estimation, especially
under lighting variations. Moreover, the lack of leveraging
differentiable rendering techniques limits their ability to perform
efficient gradient-based optimization for pose estimation.

\subsection{NeRF-Based Localization}\label{nerf-based-localization}

The advent of Neural Radiance Fields (NeRF)
\autocite{mildenhallNeRFRepresentingScenes2022} has revolutionized novel
view synthesis by representing scenes as continuous volumetric functions
learned from images. NeRF has inspired new approaches to camera
localization by leveraging its differentiable rendering capabilities.

\textbf{Pose Estimation with NeRF} involves inverting a pre-trained NeRF
model to recover camera poses by minimizing the photometric error
between rendered images and observed images. iNeRF
\autocite{yen-chenInerfInvertingNeural2021} formulates pose estimation
as an optimization problem, using gradient-based methods to refine
camera parameters. While iNeRF achieves impressive accuracy, it suffers
from high computational costs due to the per-pixel ray marching required
in NeRF's volumetric rendering pipeline. This limitation hampers its
applicability in real-time localization tasks.

\textbf{Accelerated NeRF Variants} aim to address computational
inefficiency by introducing explicit data structures. Instant-NGP
\autocite{mullerInstantNeuralGraphics2022} uses hash maps to accelerate
training and rendering, achieving interactive frame rates. PlenOctrees
\autocite{yuPlenoctreesRealtimeRendering2021} and Plenoxels
\autocite{fridovich-keilPlenoxelsRadianceFields2022} employ sparse voxel
grids to represent the scene, significantly reducing computation time.
However, even with these optimizations, rendering speeds may still not
meet the demands of real-time localization in dynamic environments.

Furthermore, NeRF-based localization methods rely heavily on photometric
consistency, making them sensitive to lighting variations, dynamic
objects, and non-Lambertian surfaces. This reliance on RGB data can
reduce robustness in real-world conditions where lighting can change
dramatically. Additionally, the extensive training required on specific
scenes limits their adaptability to new or changing environments.

\subsection{Gaussian-Based
Localization}\label{gaussian-based-localization}

Recent advancements in scene representation have introduced 3D Gaussian
splatting as an efficient alternative to NeRF. \textbf{3D Gaussian
Splatting} \autocite{kerbl3DGaussianSplatting2023} represents scenes
using a set of 3D Gaussian primitives and employs rasterization-based
rendering, offering significant computational advantages over volumetric
rendering.

\textbf{Gaussian Splatting in Localization} has been explored in methods
such as SplaTAM \autocite{keethaSplaTAMSplatTrack2024}, CG-SLAM
\autocite{huCGSLAMEfficientDense2024}, RTG-SLAM
\autocite{pengRTGSLAMRealtime3D2024}, and GS-ICP-SLAM
\autocite{haRGBDGSICPSLAM2024}. SplaTAM introduces a SLAM system that
uses gradient-based optimization to refine both the map and camera
poses, utilizing RGB-D data and 3D Gaussians for dense mapping. CG-SLAM
focuses on an uncertainty-aware 3D Gaussian field to improve tracking
and mapping performance, incorporating depth uncertainty modeling.

Pose estimation approaches in these methods often rely on traditional
point cloud registration techniques, such as Iterative Closest Point
(ICP) algorithms \autocite{beslMethodRegistration3shapes1992}.
\textbf{RTG-SLAM}\autocite{pengRTGSLAMRealtime3D2024} employs ICP for
pose estimation within a 3D Gaussian splatting framework, demonstrating
real-time performance in 3D reconstruction tasks. Similarly,
\textbf{GS-ICP-SLAM} utilizes Generalized ICP
\autocite{segalGeneralizedicp2009a} for alignment, effectively handling
the variability in point cloud density and improving robustness.

\textbf{Gaussian-SLAM}
\autocite{yugayGaussianSLAMPhotorealisticDense2024} adapts traditional
RGB-D odometry methods, combining colored point cloud alignment
\autocite{parkColoredPointCloud2017} with an energy-based visual
odometry approach \autocite{steinbruckerRealtimeVisualOdometry2011}.
These methods integrate ICP-based techniques within Gaussian-based
representations to estimate camera poses.

While effective in certain scenarios, the reliance on ICP-based methods
introduces limitations\autocite{pomerleauComparingICPVariants2013}. ICP
algorithms require good initial alignment and can be sensitive to local
minima, often necessitating careful initialization to ensure
convergence. Additionally, ICP can be computationally intensive,
especially with large point sets, hindering real-time performance. These
methods may not fully exploit the differentiable rendering capabilities
of 3D Gaussian representations for pose optimization.

Optimizing camera poses using depth information within a differentiable
rendering framework offers several advantages, particularly in
environments with challenging lighting or low-texture surfaces. Depth
data provides direct geometric information about the scene, which is
invariant to illumination changes, leading to more robust pose
estimation. By focusing on depth-only optimization, methods can achieve
robustness to lighting variations and improve computational efficiency
by avoiding the processing of color data.

However, existing Gaussian-based localization techniques have not fully
exploited depth-only optimization within a differentiable rendering
framework. Challenges such as sensor noise, incomplete depth data due to
occlusions, and the need for accurate initial pose estimates remain.
Furthermore, many approaches tightly couple mapping and localization,
introducing unnecessary computational overhead and complexity when the
primary goal is pose estimation.

These limitations motivate the development of our proposed method. By
leveraging the differentiable rendering capabilities of 3D Gaussian
splatting specifically for depth-only pose optimization, we aim to
overcome the challenges faced by existing methods. Our approach
eliminates reliance on photometric data, enhancing robustness to
lighting variations and reducing computational overhead. By decoupling
localization from mapping, we simplify the optimization process, making
it more suitable for real-time applications. Additionally, using
quaternions for rotation parameterization
\autocite{kuipersQuaternionsRotationSequences1999} and careful
initialization strategies improves the stability and convergence of the
optimization, addressing challenges associated with sensor noise and
incomplete data.

Our method fully exploits the differentiable rendering pipeline to
perform efficient gradient-based optimization for pose estimation,
setting it apart from ICP-based approaches. By focusing on depth
information and leveraging the strengths of 3D Gaussian splatting, we
provide a robust and computationally efficient solution for camera
localization in complex environments.

\section{Method}\label{method}

\begin{figure*}[!t]
\vspace{-2cm}
\centering
\resizebox{\textwidth}{!}{\includegraphics{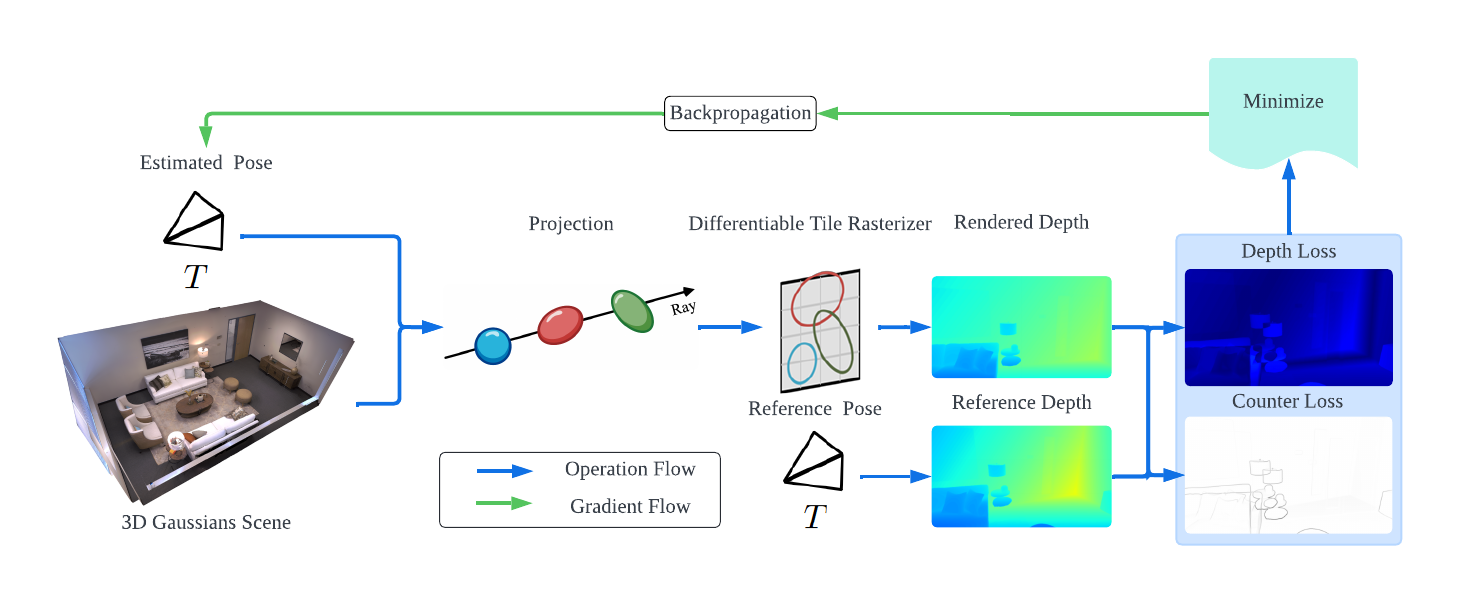}}
\caption{We propose \textbf{GSplatLoc}, a novel camera localization method that leverages the differentiable rendering capabilities of 3D Gaussian splatting for efficient and accurate pose estimation.}
\label{fig:cross-column-image}
\end{figure*}

\textbf{Overview.} We propose \textbf{GSplatLoc}, a novel camera
localization method that leverages the differentiable rendering
capabilities of 3D Gaussian splatting for efficient and accurate pose
estimation. By formulating pose estimation as a gradient-based
optimization problem within a fully differentiable framework, GSplatLoc
enables direct optimization of camera poses using depth information
rendered from a pre-existing 3D Gaussian scene representation. This
approach allows us to achieve high-precision localization suitable for
real-time applications.

\textbf{Motivation.} Traditional SLAM systems that use point clouds,
meshes, or surfels for 3D representation often face limitations in
rendering quality and computational efficiency, hindering their ability
to provide photorealistic scene exploration and fine-grained map
updates. Neural Radiance Fields (NeRF)
\autocite{mildenhallNeRFRepresentingScenes2022} have demonstrated
exceptional rendering quality but suffer from computational
inefficiencies due to per-pixel ray marching in volume rendering, making
real-time applications challenging.

The recent development of \textbf{3D Gaussian Splatting}
\autocite{kerbl3DGaussianSplatting2023} offers a promising alternative
by employing a rasterization-based rendering pipeline. In this method,
scenes are represented using a set of 3D Gaussians, which can be
efficiently projected onto the image plane and rasterized to produce
high-quality renderings at interactive frame rates. The differentiable
nature of this rendering process enables gradient computation with
respect to both the scene parameters and the camera pose.

By leveraging these properties, we aim to develop a localization method
that fully utilizes the differentiable rendering capabilities of 3D
Gaussian splatting. Our approach focuses on optimizing the camera pose
by minimizing the difference between the rendered depth map and the
observed query depth image, thus enabling accurate and efficient pose
estimation suitable for real-time SLAM systems.

\textbf{Problem Formulation.} Our objective is to estimate the 6-DoF
pose \((\mathbf{R}, \mathbf{t}) \in SE(3)\) of a query depth image
\(D_q\), where \(\mathbf{R}\) is the rotation matrix and \(\mathbf{t}\)
is the translation vector in the camera coordinate system. Given a 3D
representation of the environment in the form of 3D Gaussians, let
\(\mathcal{G} = \{G_i\}_{i=1}^N\) denote a set of \(N\) Gaussians, and
posed reference depth images \(\{D_k\}\), which together constitute the
reference data.

\subsection{Scene Representation}\label{scene-representation}

Building upon the Gaussian splatting method
\autocite{kerbl3DGaussianSplatting2023}, we adapt the scene
representation to focus on the differentiable depth rendering process,
which is crucial for our localization task. Our approach utilizes the
efficiency and quality of Gaussian splatting while tailoring it
specifically for depth-based localization.

\textbf{3D Gaussians.} Each Gaussian \(G_i\) is characterized by its 3D
mean \(\boldsymbol{\mu}_i \in \mathbb{R}^3\), 3D covariance matrix
\(\boldsymbol{\Sigma}_i \in \mathbb{R}^{3\times3}\), opacity
\(o_i \in \mathbb{R}\), and scale \(\mathbf{s}_i \in \mathbb{R}^3\). To
represent the orientation of each Gaussian, we use a rotation quaternion
\(\mathbf{q}_i \in \mathbb{R}^4\).

The 3D covariance matrix \(\boldsymbol{\Sigma}_i\) is parameterized
using \(\mathbf{s}_i\) and \(\mathbf{q}_i\):

\[
\boldsymbol{\Sigma}_i = \mathbf{R}(\mathbf{q}_i) \mathbf{S}(\mathbf{s}_i) \mathbf{S}(\mathbf{s}_i)^\top \mathbf{R}(\mathbf{q}_i)^\top
\]

where \(\mathbf{R}(\mathbf{q}_i)\) is the rotation matrix derived from
\(\mathbf{q}_i\), and
\(\mathbf{S}(\mathbf{s}_i) = \text{diag}(\mathbf{s}_i)\) is a diagonal
matrix of scales.

\textbf{Projecting 3D to 2D.} For the projection of 3D Gaussians onto
the 2D image plane, we follow the approach described by
\autocite{kerbl3DGaussianSplatting2023}. The 3D mean
\(\boldsymbol{\mu}_i\) is first transformed into the camera coordinate
frame using the world-to-camera transformation
\(\mathbf{T}_{wc} \in SE(3)\). Then, it is projected using the
projection matrix \(\mathbf{P} \in \mathbb{R}^{4 \times 4}\) and mapped
to pixel coordinates via the function
\(\pi: \mathbb{R}^4 \rightarrow \mathbb{R}^2\):

\[
\boldsymbol{\mu}_{I,i} = \pi\left( \mathbf{P} \mathbf{T}_{wc} \boldsymbol{\mu}_{i,\text{homogeneous}} \right)
\]

Similarly, the 2D covariance
\(\boldsymbol{\Sigma}_{I,i} \in \mathbb{R}^{2\times2}\) of the projected
Gaussian is obtained by transforming the 3D covariance
\(\boldsymbol{\Sigma}_i\) into the image plane: \[
\boldsymbol{\Sigma}_{I,i} = \mathbf{J} \mathbf{R}_{wc} \boldsymbol{\Sigma}_i \mathbf{R}_{wc}^\top \mathbf{J}^\top
\]

where \(\mathbf{R}_{wc}\) represents the rotation component of
\(\mathbf{T}_{wc}\), and \(\mathbf{J}\) is the Jacobian of the
projection function, accounting for the affine transformation from 3D to
2D as described by \autocite{zwickerEWASplatting2002}.

\subsection{Depth Rendering}\label{depth-rendering}

We implement a differentiable depth rendering process, which is crucial
for our localization method as it allows for gradient computation
throughout the rendering pipeline. This differentiability enables us to
optimize camera poses directly based on rendered depth maps.

\textbf{Compositing Depth.} For depth map generation, we employ a
front-to-back compositing scheme, which allows for accurate depth
estimation and proper handling of occlusions. Let \(d_n\) denote the
depth value of the \(n\)-th Gaussian, corresponding to the z-coordinate
of its mean in the camera coordinate system. The depth at pixel
\(\mathbf{p}\), denoted \(D(\mathbf{p})\), is computed as
\autocite{kerbl3DGaussianSplatting2023}:

\[D(\mathbf{p}) = \sum_{n \leq N} d_n \cdot \alpha_n \cdot T_n,\]

where \(T_n\) is the cumulative transparency up to the \((n-1)\)-th
Gaussian:

\[T_n = \prod_{m < n} (1 - \alpha_m).\]

In this formulation, \(\alpha_n\) represents the opacity contribution of
the \(n\)-th Gaussian at pixel \(\mathbf{p}\), calculated as:

\[\alpha_n = o_n \cdot \exp(-\sigma_n),\]

with

\[\sigma_n = \frac{1}{2} \boldsymbol{\Delta}_n^\top \boldsymbol{\Sigma}_I^{-1} \boldsymbol{\Delta}_n.\]

Here, \(\boldsymbol{\Delta}_n\) is the offset between the pixel center
and the center of the 2D Gaussian \(\boldsymbol{\mu}_{I,i}\), and
\(\boldsymbol{\Sigma}_I\) is the 2D covariance of the projected
Gaussian. The opacity parameter \(o_n\) controls the overall opacity of
the Gaussian.

\textbf{Normalization of Depth.} To ensure consistent depth
representation across the image, we normalize the accumulated depth
values. We first compute the total accumulated opacity at each pixel
\(\mathbf{p}\): \[
\alpha(\mathbf{p}) = \sum_{n \leq N} \alpha_n \cdot T_n
\]

The normalized depth at pixel \(\mathbf{p}\) is then defined as:

\[
\text{Norm}_D(\mathbf{p}) = \frac{D(\mathbf{p})}{\alpha(\mathbf{p})}
\]

This normalization ensures that the depth values are properly scaled,
making them comparable across different regions of the image, even when
the density of Gaussians varies.

The differentiable nature of this depth rendering process is key to our
localization method. It allows us to compute gradients with respect to
the Gaussian parameters and camera pose, enabling direct optimization
based on the rendered depth maps. This differentiability facilitates
efficient gradient-based optimization, forming the foundation for our
subsequent localization algorithm.

\subsection{Localization as Image
Alignment}\label{localization-as-image-alignment}

Assuming we have an existing map represented by a set of 3D Gaussians,
our localization task focuses on estimating the 6-DoF pose of a query
depth image \(D_q\) within this map. This process essentially becomes an
image alignment problem between the rendered depth map from our Gaussian
representation and the query depth image.

\textbf{Rotating with
Quaternions.}\autocite{kuipersQuaternionsRotationSequences1999} We
parameterize the camera pose using a quaternion \(\mathbf{q}_{cw}\) for
rotation and a vector \(\mathbf{t}_{cw}\) for translation. This choice
of parameterization is particularly advantageous in our differential
rendering context. Quaternions provide a continuous and singularity-free
representation of rotation, which is crucial for gradient-based
optimization. Moreover, their compact four-parameter form aligns well
with our differentiable rendering pipeline, allowing for efficient
computation of gradients with respect to rotation parameters.

\textbf{Loss Function.} Our optimization strategy leverages the
differentiable nature of our depth rendering process. We define a loss
function that incorporates both depth accuracy and edge alignment:

\[
\mathcal{L} = \lambda_1 \mathcal{L}_d + \lambda_2 \mathcal{L}_c
\]

where \(\lambda_1\) and \(\lambda_2\) are weighting factors (typically
set to 0.8 and 0.2, respectively) that balance the contributions of the
depth and contour losses. The depth loss \(\mathcal{L}_d\) measures the
L1 difference between the rendered depth map and the observed depth
image:

\[
\mathcal{L}_d = \sum_{i \in \mathcal{M}} \left| D_i^{\text{rendered}} - D_i^{\text{observed}} \right|
\]

The contour loss \(\mathcal{L}_c\) focuses on aligning the depth
gradients (edges) between the rendered and observed depth images:

\[
\mathcal{L}_c = \sum_{j \in \mathcal{M}} \left| \nabla D_j^{\text{rendered}} - \nabla D_j^{\text{observed}} \right|
\]

Here, \(\nabla D\) represents the gradient of the depth image, computed
using the Sobel operator \autocite{kanopoulosDesignImageEdge1988}, and
\(\mathcal{M}\) is the mask of valid pixels determined by the rendered
alpha mask.

The contour loss \(\mathcal{L}_{c}\) serves several crucial purposes. It
ensures that depth discontinuities in the rendered image align well with
those in the observed depth image, thereby improving the overall
accuracy of the pose estimation. By explicitly considering edge
information, we preserve important structural features of the scene
during optimization. Furthermore, the contour loss is less sensitive to
absolute depth values and more focused on relative depth changes, making
it robust to global depth scale differences.

The optimization objective can be formulated as:

\[
\min_{\mathbf{q}_{cw}, \mathbf{t}_{cw}} \mathcal{L} + \lambda_q \|\mathbf{q}_{cw}\|_2^2 + \lambda_t \|\mathbf{t}_{cw}\|_2^2
\]

where \(\lambda_q\) and \(\lambda_t\) are regularization coefficients
for the quaternion and translation parameters, respectively.

\textbf{Masking Uncertainty.} The rendered alpha mask plays a crucial
role in our optimization process. It effectively captures the epistemic
uncertainty of our map, allowing us to focus the optimization on
well-represented parts of the scene. By utilizing this mask, we avoid
optimizing based on unreliable or non-existent data, which could
otherwise lead to erroneous pose estimates.

\textbf{Optimization Parameters.} We perform the optimization using the
Adam optimizer, with distinct learning rates and weight decay values for
the rotation and translation parameters. Specifically, we set the
learning rate for quaternion optimization to \(5 \times 10^{-4}\) and
for translation optimization to \(10^{-3}\), based on empirical tuning.
Weight decay, set to \(10^{-3}\) for both parameters, acts as a
regularization term to prevent overfitting. These settings balance the
trade-off between convergence speed and optimization stability.

\subsection{Pipeline}\label{pipeline}

Our GSplatLoc method streamlines the localization process by utilizing
only the posed reference depth images \(\{D_k\}\) and the query depth
image \(D_q\). The differentiable rendering of 3D Gaussians enables
efficient and smooth convergence during optimization.

\textbf{Evaluation Scene.} For consistent evaluation, we initialize the
3D Gaussians from point clouds derived from the posed reference depth
images \(\{D_k\}\). Each point in the point cloud corresponds to the
mean \(\boldsymbol{\mu}_i\) of a Gaussian \(G_i\). After filtering out
outliers, we set the opacity \(o_i = 1\) for all Gaussians to ensure
full contribution in rendering. The scale \(\mathbf{s}_i\) is
initialized isotropically based on the local point density:

\[
\mathbf{s}_i = (\sigma_i, \sigma_i, \sigma_i), \quad \text{with } \sigma_i = \sqrt{\frac{1}{3}\sum_{j=1}^3 d_{ij}^2}
\]

where \(d_{ij}\) is the distance to the \(j\)-th nearest neighbor of
point \(i\), computed using k-nearest neighbors (with \(k=4\)). This
initialization balances the representation of local geometry. The
rotation quaternion \(\mathbf{q}_i\) is initially set to
\((1, 0, 0, 0)\) for all Gaussians, corresponding to no rotation.

To further enhance optimization stability, we apply standard Principal
Component Analysis (PCA) to align the principal axes of the point cloud
with the coordinate axes. By centering the point cloud at its mean and
aligning its principal axes, we normalize the overall scene orientation.
This provides a more uniform starting point for optimization across
diverse datasets, significantly improving the stability of the loss
reduction during optimization and facilitating the attainment of lower
final loss values, especially in the depth loss component of our
objective function.

\textbf{Optimization.} We employ the
Adam\autocite{kingmaAdamMethodStochastic2014} optimizer for optimizing
both the quaternion and translation parameters, using the distinct
learning rates and weight decay values as previously described. The
optimization process greatly benefits from the real-time rendering
capabilities of 3D Gaussian splatting. Since rendering is extremely
fast, each iteration of the optimizer is limited mainly by the rendering
speed, allowing for rapid convergence of our pose estimation algorithm
and making it suitable for real-time applications. Our optimization
approach consistently achieves sub-millimeter accuracy (average ATE RMSE
of \textbf{0.01587 cm}) on synthetic datasets, while maintaining robust
performance in real-world scenarios.

\textbf{Convergence.} To determine convergence, we implement an early
stopping mechanism based on the stabilization of the total loss. Our
experiments show that the total loss usually stabilizes after
approximately 100 iterations. We employ a patience parameter: after 100
iterations, if the total loss does not decrease for a predefined number
of consecutive iterations, the optimization loop is terminated. We then
select the pose estimate corresponding to the minimum total loss as the
optimal pose.

In summary, our pipeline effectively combines the efficiency of Gaussian
splatting with a robust optimization strategy, resulting in a fast and
accurate camera localization method suitable for real-time applications.

\section{Evaluation}\label{evaluation}

We conducted a comprehensive evaluation spanning both synthetic and
real-world environments, with pose estimation errors ranging from as low
as \textbf{0.01587 cm} in controlled settings to competitive performance
(\textbf{0.80982 cm}) in challenging real-world scenarios. Our
evaluation framework encompasses multiple aspects of localization
performance, from implementation details to dataset selection and
baseline comparisons.

\subsection{Experimental Setup}\label{experimental-setup}

\textbf{Implementation Details.} Our localization pipeline was
implemented on a system equipped with an Intel Core i7-13620H CPU,
16\,GB of RAM, and an NVIDIA RTX 4060 GPU with 8\,GB of memory. The
algorithm was developed using Python and PyTorch, utilizing custom CUDA
kernels to accelerate the rasterization and backpropagation processes
inherent in our differentiable rendering approach. This setup ensures
that our method achieves real-time performance, which is crucial for
practical applications in SLAM systems.

\textbf{Datasets.} We evaluated our method on two widely recognized
datasets for SLAM benchmarking: the \textbf{Replica} dataset
\autocite{straubReplicaDatasetDigital2019} and the \textbf{TUM RGB-D}
dataset \autocite{sturmBenchmarkEvaluationRGBD2012}. The Replica dataset
provides high-fidelity synthetic indoor environments, ideal for
controlled evaluations of localization algorithms. We utilized data
collected by Sucar et al. \autocite{sucarImapImplicitMapping2021}, which
includes trajectories from an RGB-D sensor with ground-truth poses. The
TUM RGB-D dataset\autocite{sturmBenchmarkEvaluationRGBD2012} offers
real-world sequences captured in various indoor settings, providing a
diverse range of scenarios to test the robustness of our method.

\textbf{Metrics.} Localization accuracy was assessed using two standard
metrics: the \textbf{Absolute Trajectory Error (ATE RMSE)}, measured in
centimeters, and the \textbf{Absolute Angular Error (AAE RMSE)},
measured in degrees. The ATE RMSE quantifies the root mean square error
between the estimated and ground-truth camera positions, while the AAE
RMSE measures the accuracy of the estimated camera orientations.

\textbf{Baselines.}~To provide a comprehensive comparison, we evaluated
our method against several state-of-the-art SLAM systems that leverage
advanced scene representations. Specifically, we compared against
RTG-SLAM(ICP)\autocite{pengRTGSLAMRealtime3D2024}, which utilizes
Iterative Closest Point (ICP) for pose estimation within a 3D Gaussian
splatting framework. We also included GS-ICP-SLAM(GICP)
\autocite{haRGBDGSICPSLAM2024}, which employs Generalized ICP for
alignment in a Gaussian-based representation. Additionally, we
considered Gaussian-SLAM
\autocite{yugayGaussianSLAMPhotorealisticDense2024}, evaluating both its
PLANE ICP and HYBRID variants, which adapt traditional RGB-D odometry
methods by incorporating plane-based ICP and a hybrid approach combining
photometric and geometric information. These baselines were selected
because they represent the current state of the art in SLAM systems
utilizing advanced scene representations and focus on the localization
component, which aligns with the scope of our work.

\subsection{Localization Evaluation}\label{localization-evaluation}

We conducted comprehensive experiments on both synthetic and real-world
datasets to evaluate the performance of GSplatLoc against
state-of-the-art methods utilizing advanced scene representations.

\begin{table}[htbp]
\renewcommand{\thetable}{\textbf{\arabic{table}}}
\renewcommand{\tablename}{\textbf{Table}}
\centering
\caption{\textbf{Replica\cite{straubReplicaDatasetDigital2019} (ATE RMSE ↓[cm]).}}
\label{table:_textbf_replica_cite}
\begin{adjustbox}{max width=\columnwidth,max height=!,center}
\begin{tabular}{lccccccccc}
\toprule
\textbf{Methods} & \textbf{Avg.} & \textbf{R0} & \textbf{R1} & \textbf{R2} & \textbf{Of0} & \textbf{Of1} & \textbf{Of2} & \textbf{Of3} & \textbf{Of4}\\
\midrule
RTG-SLAM(ICP)\cite{pengRTGSLAMRealtime3D2024} & 1.102 & 1.286 & 0.935 & \cellcolor{yellow!30}1.117 & 0.983 & 0.626 & 1.194 & \cellcolor{yellow!30}1.334 & 1.340\\
GS-ICP-SLAM(GICP)\cite{haRGBDGSICPSLAM2024} & \cellcolor{yellow!30}1.084 & 1.250 & \cellcolor{yellow!30}0.828 & 1.183 & \cellcolor{lime!50}0.924 & \cellcolor{lime!50}0.591 & \cellcolor{lime!50}1.175 & 1.438 & \cellcolor{yellow!30}1.284\\
Gaussian-SLAM(PLANE ICP)\cite{yugayGaussianSLAMPhotorealisticDense2024} & \cellcolor{lime!50}1.086 & \cellcolor{yellow!30}1.246 & 0.855 & 1.186 & \cellcolor{yellow!30}0.922 & \cellcolor{yellow!30}0.590 & \cellcolor{yellow!30}1.162 & \cellcolor{lime!50}1.426 & 1.304\\
Gaussian-SLAM(HYBRID)\cite{yugayGaussianSLAMPhotorealisticDense2024} & 1.096 & \cellcolor{lime!50}1.248 & \cellcolor{lime!50}0.831 & \cellcolor{lime!50}1.183 & 0.926 & 0.595 & 1.201 & 1.499 & \cellcolor{lime!50}1.289\\
\midrule
\textbf{Ours} & \cellcolor{green!30}\textbf{0.016} & \cellcolor{green!30}\textbf{0.015} & \cellcolor{green!30}\textbf{0.013} & \cellcolor{green!30}\textbf{0.021} & \cellcolor{green!30}\textbf{0.011} & \cellcolor{green!30}\textbf{0.009} & \cellcolor{green!30}\textbf{0.018} & \cellcolor{green!30}\textbf{0.020} & \cellcolor{green!30}\textbf{0.019}\\
\bottomrule
\end{tabular}
\end{adjustbox}
\end{table}

\textbf{Table 1.} presents the Absolute Trajectory Error (ATE RMSE)
results on the Replica dataset. Our method achieves remarkable
performance with an average ATE RMSE of \textbf{0.01587 cm},
significantly outperforming existing approaches by nearly two orders of
magnitude. The closest competitor, RTG-SLAM(ICP)
\autocite{pengRTGSLAMRealtime3D2024}, achieves an average error of
1.10186 cm. This substantial improvement is consistent across all
sequences, with particularly notable performance in challenging scenes
like Of1 (0.00937 cm) and R1 (0.01272 cm).

\begin{table}[htbp]
\renewcommand{\thetable}{\textbf{\arabic{table}}}
\renewcommand{\tablename}{\textbf{Table}}
\centering
\caption{\textbf{Replica\cite{straubReplicaDatasetDigital2019} (AAE RMSE ↓[°]).}}
\label{table:_textbf_replica_cite}
\begin{adjustbox}{max width=\columnwidth,max height=!,center}
\begin{tabular}{lccccccccc}
\toprule
\textbf{Methods} & \textbf{Avg.} & \textbf{R0} & \textbf{R1} & \textbf{R2} & \textbf{Of0} & \textbf{Of1} & \textbf{Of2} & \textbf{Of3} & \textbf{Of4}\\
\midrule
RTG-SLAM(ICP)\cite{pengRTGSLAMRealtime3D2024} & \cellcolor{yellow!30}0.471 & \cellcolor{yellow!30}0.429 & \cellcolor{yellow!30}0.690 & \cellcolor{yellow!30}0.544 & \cellcolor{yellow!30}0.640 & \cellcolor{yellow!30}0.336 & \cellcolor{yellow!30}0.434 & \cellcolor{yellow!30}0.281 & \cellcolor{yellow!30}0.419\\
GS-ICP-SLAM(GICP)\cite{haRGBDGSICPSLAM2024} & 0.631 & 0.476 & 0.812 & 0.781 & 0.709 & 0.537 & 0.662 & 0.446 & 0.624\\
Gaussian-SLAM(PLANE ICP)\cite{yugayGaussianSLAMPhotorealisticDense2024} & \cellcolor{lime!50}0.593 & \cellcolor{lime!50}0.465 & \cellcolor{lime!50}0.772 & \cellcolor{lime!50}0.723 & \cellcolor{lime!50}0.681 & \cellcolor{lime!50}0.522 & \cellcolor{lime!50}0.582 & \cellcolor{lime!50}0.438 & \cellcolor{lime!50}0.558\\
Gaussian-SLAM(HYBRID)\cite{yugayGaussianSLAMPhotorealisticDense2024} & 0.633 & 0.476 & 0.812 & 0.781 & 0.709 & 0.541 & 0.667 & 0.449 & 0.625\\
\midrule
\textbf{Ours} & \cellcolor{green!30}\textbf{0.009} & \cellcolor{green!30}\textbf{0.007} & \cellcolor{green!30}\textbf{0.008} & \cellcolor{green!30}\textbf{0.010} & \cellcolor{green!30}\textbf{0.009} & \cellcolor{green!30}\textbf{0.009} & \cellcolor{green!30}\textbf{0.011} & \cellcolor{green!30}\textbf{0.009} & \cellcolor{green!30}\textbf{0.011}\\
\bottomrule
\end{tabular}
\end{adjustbox}
\end{table}

\textbf{Table 2.} GSplatLoc achieves an average AAE RMSE of
\textbf{0.00925°}. This represents a significant improvement over
traditional ICP-based methods, with
RTG-SLAM\autocite{pengRTGSLAMRealtime3D2024} and
GS-ICP-SLAM\autocite{haRGBDGSICPSLAM2024} showing average errors of
0.47141° and 0.63100° respectively. The performance advantage is
particularly evident in sequences with complex rotational movements,
such as Of2 and Of4, where our method maintains sub-0.01° accuracy.

\begin{table}[htbp]
\renewcommand{\thetable}{\textbf{\arabic{table}}}
\renewcommand{\tablename}{\textbf{Table}}
\centering
\caption{\textbf{TUM\cite{sturmBenchmarkEvaluationRGBD2012} (ATE RMSE ↓[cm]).}}
\label{table:_textbf_tum_cite_stu}
\begin{adjustbox}{max width=\columnwidth,max height=!,center}
\begin{tabular}{lcccccc}
\toprule
\textbf{Methods} & \textbf{Avg.} & \textbf{fr1/desk} & \textbf{fr1/desk2} & \textbf{fr1/room} & \textbf{fr2/xyz} & \textbf{fr3/off.}\\
\midrule
RTG-SLAM(ICP)\cite{pengRTGSLAMRealtime3D2024} & \cellcolor{green!30}\textbf{0.576} & \cellcolor{green!30}\textbf{0.720} & \cellcolor{green!30}\textbf{0.826} & \cellcolor{yellow!30}0.744 & \cellcolor{green!30}\textbf{0.054} & \cellcolor{green!30}\textbf{0.537}\\
GS-ICP-SLAM(GICP)\cite{haRGBDGSICPSLAM2024} & 1.955 & 2.265 & 3.493 & 2.783 & 0.287 & \cellcolor{lime!50}0.945\\
Gaussian-SLAM(PLANE ICP)\cite{yugayGaussianSLAMPhotorealisticDense2024} & \cellcolor{lime!50}1.279 & \cellcolor{lime!50}1.659 & 1.951 & 1.607 & \cellcolor{lime!50}0.281 & \cellcolor{yellow!30}0.895\\
Gaussian-SLAM(HYBRID)\cite{yugayGaussianSLAMPhotorealisticDense2024} & 1.287 & 1.834 & \cellcolor{lime!50}1.880 & \cellcolor{lime!50}1.398 & 0.305 & 1.019\\
\midrule
\textbf{Ours} & \cellcolor{yellow!30}0.810 & \cellcolor{yellow!30}0.931 & \cellcolor{yellow!30}1.006 & \cellcolor{green!30}\textbf{0.666} & \cellcolor{yellow!30}0.248 & 1.197\\
\bottomrule
\end{tabular}
\end{adjustbox}
\end{table}

\textbf{Table 3.} presents the ATE RMSE in centimeters for various
methods on the TUM-RGBD dataset . Our method achieves competitive
results with an average ATE RMSE of \textbf{8.0982 cm}, outperforming
GS-ICP-SLAM\autocite{haRGBDGSICPSLAM2024} and
Gaussian-SLAM\autocite{yugayGaussianSLAMPhotorealisticDense2024} in most
sequences. While RTG-SLAM\autocite{pengRTGSLAMRealtime3D2024} shows
lower errors in some sequences, our method consistently provides
accurate pose estimates across different environments. The increased
error compared to the Replica dataset is expected due to the real-world
challenges present in the TUM RGB-D
dataset\autocite{sturmBenchmarkEvaluationRGBD2012}, such as sensor noise
and environmental variability. Despite these challenges, our method
demonstrates robustness and maintains reasonable localization accuracy.

\textbf{Tables 3.} presents results on the more challenging TUM RGB-D
dataset\autocite{sturmBenchmarkEvaluationRGBD2012}, which introduces
real-world complexities such as sensor noise and dynamic environments.
In terms of translational accuracy, GSplatLoc achieves competitive
performance with an average ATE RMSE of \textbf{0.80982 cm}. While
RTG-SLAM\autocite{pengRTGSLAMRealtime3D2024} shows slightly better
average performance (0.57636 cm), our method consistently outperforms
both GS-ICP-SLAM\autocite{haRGBDGSICPSLAM2024} (1.95454 cm) and
Gaussian-SLAM\autocite{yugayGaussianSLAMPhotorealisticDense2024}
variants (1.27873 cm and 1.28716 cm) across most sequences.

\begin{table}[htbp]
\renewcommand{\thetable}{\textbf{\arabic{table}}}
\renewcommand{\tablename}{\textbf{Table}}
\centering
\caption{\textbf{TUM\cite{sturmBenchmarkEvaluationRGBD2012} (AAE RMSE ↓[°]).}}
\label{table:_textbf_tum_cite_stu}
\begin{adjustbox}{max width=\columnwidth,max height=!,center}
\begin{tabular}{lcccccc}
\toprule
\textbf{Methods} & \textbf{Avg.} & \textbf{fr1/desk} & \textbf{fr1/desk2} & \textbf{fr1/room} & \textbf{fr2/xyz} & \textbf{fr3/off.}\\
\midrule
RTG-SLAM(ICP)\cite{pengRTGSLAMRealtime3D2024} & \cellcolor{green!30}\textbf{0.916} & \cellcolor{yellow!30}1.181 & \cellcolor{yellow!30}1.557 & \cellcolor{yellow!30}1.355 & \cellcolor{yellow!30}0.138 & \cellcolor{green!30}\textbf{0.347}\\
GS-ICP-SLAM(GICP)\cite{haRGBDGSICPSLAM2024} & 1.117 & 1.426 & 2.098 & 1.594 & \cellcolor{green!30}\textbf{0.114} & \cellcolor{yellow!30}0.355\\
Gaussian-SLAM(PLANE ICP)\cite{yugayGaussianSLAMPhotorealisticDense2024} & \cellcolor{yellow!30}0.959 & \cellcolor{lime!50}1.288 & \cellcolor{lime!50}1.618 & \cellcolor{lime!50}1.363 & \cellcolor{lime!50}0.147 & \cellcolor{lime!50}0.381\\
Gaussian-SLAM(HYBRID)\cite{yugayGaussianSLAMPhotorealisticDense2024} & 1.090 & 1.388 & 1.791 & 1.564 & 0.182 & 0.525\\
\midrule
\textbf{Ours} & \cellcolor{lime!50}0.979 & \cellcolor{green!30}\textbf{1.126} & \cellcolor{green!30}\textbf{1.265} & \cellcolor{green!30}\textbf{0.907} & 0.789 & 0.808\\
\bottomrule
\end{tabular}
\end{adjustbox}
\end{table}

\textbf{Table 4.} The rotational accuracy results on TUM RGB-D
dataset\autocite{sturmBenchmarkEvaluationRGBD2012} demonstrate the
robustness of our approach in real-world scenarios. GSplatLoc maintains
stable performance with an average AAE RMSE of \textbf{0.97928°},
comparable to RTG-SLAM's 0.91561°. Notably, our method shows superior
performance in challenging sequences like fr1/room (0.90722°) compared
to competing methods, which exhibit errors ranging from 1.35470° to
1.56370°.

The performance gap between synthetic and real-world results highlights
the impact of sensor noise and environmental complexity on localization
accuracy. While the near-perfect accuracy achieved on the Replica
dataset demonstrates the theoretical capabilities of our approach, the
competitive performance on TUM RGB-D
dataset\autocite{sturmBenchmarkEvaluationRGBD2012} validates its
practical applicability in real-world scenarios.

\subsection{Discussion}\label{discussion}

The experimental results indicate that our method consistently achieves
high localization accuracy, particularly in terms of translational
error, where we significantly outperform existing approaches on the
Replica dataset. The rotational accuracy of our method is also
competitive, often surpassing other methods in challenging sequences.
These outcomes demonstrate the effectiveness of our approach in
leveraging the differentiable rendering capabilities of 3D Gaussian
splatting for pose estimation.

Several factors contribute to the superior performance of our method. By
utilizing a fully differentiable depth rendering process, our method
allows for efficient gradient-based optimization of camera poses,
leading to precise alignment between the rendered and observed depth
images. The combination of depth loss and contour loss in our
optimization objective enables the method to capture both absolute depth
differences and structural features, enhancing the robustness and
accuracy of pose estimation. Additionally, employing quaternions for
rotation representation provides a continuous and singularity-free
parameter space, improving the stability and convergence of the
optimization process.

While our method shows excellent performance on the Replica dataset, the
increased errors on the TUM RGB-D
dataset\autocite{sturmBenchmarkEvaluationRGBD2012} highlight areas for
potential improvement. Real-world datasets introduce challenges such as
sensor noise, dynamic objects, and incomplete depth data due to
occlusions. Addressing these challenges in future work could further
enhance the robustness of our method.

\subsection{Limitations}\label{limitations}

Despite the promising results, our method has certain limitations. The
reliance on accurate depth data means that performance may degrade in
environments where the depth sensor data is noisy or incomplete.
Additionally, our current implementation focuses on frame-to-frame pose
estimation with initialization from the previous frame's ground-truth
pose. In practical applications, this assumption may not hold, and
integrating our method into a full SLAM system with robust
initialization and loop closure capabilities would be necessary.
Furthermore, handling dynamic scenes and improving computational
efficiency for large-scale environments remain areas for future
exploration.

\section{Conclusion}\label{conclusion}

In this paper, we introduced \textbf{GSplatLoc}, a novel method for
ultra-precise camera localization that leverages the differentiable
rendering capabilities of 3D Gaussian splatting. By formulating pose
estimation as a gradient-based optimization problem within a fully
differentiable framework, our approach enables efficient and accurate
alignment between rendered depth maps from a pre-existing 3D Gaussian
scene and observed depth images.

Extensive experiments on the Replica and TUM RGB-D
dataset\autocite{sturmBenchmarkEvaluationRGBD2012} demonstrate that
GSplatLoc significantly outperforms state-of-the-art SLAM systems in
terms of both translational and rotational accuracy. On the Replica
dataset, our method achieves an average Absolute Trajectory Error (ATE
RMSE) of \textbf{0.01587 cm}, surpassing existing approaches by an order
of magnitude. The method also maintains competitive performance on the
TUM RGB-D dataset\autocite{sturmBenchmarkEvaluationRGBD2012}, exhibiting
robustness in real-world scenarios despite challenges such as sensor
noise and dynamic elements.

The superior performance of GSplatLoc can be attributed to several key
factors. The utilization of a fully differentiable depth rendering
process allows for efficient gradient-based optimization of camera
poses. The combination of depth and contour losses in our optimization
objective captures both absolute depth differences and structural
features, enhancing the accuracy of pose estimation. Moreover, employing
quaternions for rotation representation provides a continuous and
singularity-free parameter space, improving the stability and
convergence of the optimization process.

While the results are promising, there are limitations to address in
future work. The reliance on accurate depth data implies that
performance may degrade with noisy or incomplete sensor information.
Integrating GSplatLoc into a full SLAM system with robust
initialization, loop closure, and the capability to handle dynamic
scenes would enhance its applicability. Additionally, exploring methods
to improve computational efficiency for large-scale environments remains
an important direction for future research.

In conclusion, GSplatLoc represents a significant advancement in camera
localization accuracy for SLAM systems, setting a new standard for
localization techniques in dense mapping. The method's ability to
achieve ultra-precise pose estimation has substantial implications for
applications in robotics and augmented reality, where accurate and
efficient localization is critical.

\printbibliography
\end{document}